\documentclass[sigconf]{acmart}

\usepackage{booktabs} 

\graphicspath{{images/}}
\DeclareGraphicsExtensions{.pdf,.jpeg,.png}

\usepackage[caption=false,font=footnotesize]{subfig}

\usepackage{xspace}
\usepackage[nolist,nohyperlinks]{acronym}

\usepackage{bm}

\setcopyright{rightsretained}

\copyrightyear{2017} 
\acmYear{2017} 
\setcopyright{rightsretained} 
\acmConference[HIP2017]{The 4th International Workshop on Historical Document Imaging and Processing}{November 10--11, 2017}{Kyoto, Japan}
\acmDOI{10.1145/3151509.3151519}
\acmISBN{978-1-4503-5390-8/17/11}

\acmPrice{0.0}

\newacro{dia}[\textsc{Dia\xspace}]{Document Image Analysis}

\newacro{nln}[\textit{N-light-N}\xspace]{\textit{N-light-N}}

\newacro{scae}[\textsc{SCAE\xspace}]{Stacked Convolutional Auto-Encoder}

\newacro{ae}[\textsc{AE\xspace}]{Auto-Encoder}
\newacroplural{ae}[\textsc{AEs\xspace}]{Auto-Encoders}
\newacro{cae}[\textsc{CAE\xspace}]{Convolutional Auto-Encoder}

\newacro{nn}[\textsc{NN\xspace}]{Neural Network}
\newacroplural{nn}[\textsc{NNs\xspace}]{Neural Networks}
\newacro{ann}[\textsc{ANN\xspace}]{Artificial Neural Network}

\newacro{aenn}[\textsc{AENN\xspace}]{Auto-Encoder Neural Network}

\newacro{cnn}[\textsc{CNN\xspace}]{Convolutional Neural Network}
\newacro{ffcnn}[\textsc{FFCNN\xspace}]{Feed Forward Convolutional Neural Network}

\newacro{dnn}[\textsc{DNN\xspace}]{Deep Neural Network}
\newacroplural{dnn}[\textsc{DNNs\xspace}]{Deep Neural Networks}

\newacro{pca}[\textsc{PCA\xspace}]{Principal Component Analysis}
\newacro{lda}[\textsc{LDA\xspace}]{Linear Discriminant Analysis}

\makeatletter
\g@addto@macro\normalsize{%
  \setlength\abovedisplayskip{0pt}
  \setlength\belowdisplayskip{0pt}
  \setlength\abovedisplayshortskip{-5pt}
  \setlength\belowdisplayshortskip{4pt}
}
\makeatother

\begin{document}

\title[Hist. Doc. Image Segmentation with LDA-Init DNN]{Historical Document Image Segmentation with LDA-Initialized Deep Neural Networks}


\author{Michele Alberti, Mathias Seuret, Vinaychandran Pondenkandath, Rolf Ingold,  Marcus Liwicki}

\affiliation{%
    \department{Document Image and Voice Analysis Group (DIVA)}
    \institution{University of Fribourg}
    \streetaddress{Bd. de Perolles 90}
    \city{Fribourg} 
    \state{Switzerland} 
}
\email{{firstname}.{lastname}@unifr.ch}

\renewcommand{\shortauthors}{M. Alberti et al.}

\begin{abstract}

In this paper, we present a novel approach to perform deep neural networks layer-wise weight initialization using \ac{lda}. %
Typically, the weights of a deep neural network are initialized with: random values, greedy layer-wise pre-training (usually as Deep Belief Network or as auto-encoder) or by re-using the layers from another network (transfer learning). %
Hence, many training epochs are needed before meaningful weights are learned, or a rather similar dataset is required for seeding a fine-tuning of transfer learning. %
In this paper, we describe how to turn an \ac{lda} into either a neural layer or a classification layer.  
We analyze the initialization technique on historical documents. %
First, we show that an \ac{lda}-based initialization is quick and leads to a very stable initialization. %
Furthermore, for the task of layout analysis at pixel level, we investigate the effectiveness of \ac{lda}-based initialization and show that it outperforms state-of-the-art random weight initialization methods. %
\end{abstract}

%
%



\maketitle

\section{Introduction}

Very \ac{dnn}  are now widely used in machine learning for solving tasks in various domains. 

\par %
Although artificial neurons have been around for a long time~\cite{mcculloch1943logical}, the depth of commonly used artificial neural networks has started to increase significantly only for roughly 15 years\footnote{Note that deep neural architectures where proposed already much earlier, but they have not been heavily used in practice~\cite{888}.}\cite{mit-book}.
This is due to both: the coming back of layer-wise training methods\footnote{Referred to as Deep Belief Networks~\cite{hinton2006fast}, and often composed of Restricted Boltzmann Machines~\cite{smolensky1986}.}\cite{ballard1987modular} and the higher computational power available to researchers.

\par %
Historical \ac{dia} is an example of a domain where \ac{dnn} have been successfully applied recently.
As historical documents can be quite diverse, simple networks with few inputs usually lead to poor results, so large networks have to be used.
The diversity of the documents has several origins: different degradations (e.g ink fading or stains), complexity and variability of the layouts, overlapping of contents, writing styles, bleed-through, etc.

\par %
Because of their success, a lot of resources have been invested into research and development of \ac{dnn}. 
However, they still suffer from two major drawbacks.
The first is that, despite the computational power of new processors and \texttt{GPU}s, the training of \ac{dnn} still takes some time. 
Especially for large networks, the training time becomes a crucial issue, not only because there are more weights to use in the computations, but also because more training epochs are required for the weights to converge.
The second drawback is that initializing the weights of \ac{dnn} with random values implies that different networks will find different local error minima.

\par %
In our previous work \cite{seuretalberti2017pca} we proposed to initialize a \ac{cnn} layer-wise with \ac{pca} instead of random initialization. 
We also have shown how features which are good for one task do not necessarily generalize well to other tasks \cite{alberti2017whatYouExpectIsNOTWhatYouGet, alberti2017questioningFeatures}.
Per extension, we argue that features obtained by maximizing the variance of the input data --- which is what \ac{pca} features do --- might not be the optimal ones for performing classification tasks. 
To this end, we investigate the performances of initializing a \ac{cnn} layer-wise with a goal oriented (supervised) algorithm such as \ac{lda} by performing layout analysis at the pixel level on historical documents.

\subsection*{Contribution}

In this paper, we present a novel initialization method based on \ac{lda} which allows to quickly initialize the weights of a \ac{cnn} layer-wise\footnote{A neural layer can be both initialized to perform either features extraction (\ac{lda} space transform) or classification (\ac{lda} discriminants).} with data-based values. 
We show that such initialization is very stable\footnote{It leads to highly similar patterns of weights in networks initialized on different random samples from the same dataset.}, converge faster and to better performances when compared with the same architecture initialized with random weights. 
Additionally, even before the fine-tuning a network initialized with \ac{lda} exhibits noticeable results for classification task.



\subsection*{Related work}
\label{ssct:related_work}

Follows a brief review of literature relevant for this work.

\subsubsection*{Random Neural Network Initialization}
There are currently three main trends for neural network initialization: layer-wise unsupervised pre-training~\cite{ballard1987modular,hinton2006fast}, transfer learning~\cite{caruana1998multitask} or random initial initial weights~\cite{bottou1988reconnaissance}.
Random initialization is fast and simple to implement.
The most used approach is to initialize weights of a neuron in $\left[-\sqrt{n}, \sqrt{n}\right]$, where $n$ is the number of inputs of the neuron.

\subsubsection*{PCA Neural Network Initialization}
In our previous work \cite{seuretalberti2017pca} we successfully initialized a \ac{cnn} layer-wise with \ac{pca}.
In this work, we introduced a mathematical framework for generating \ac{cae} out of the \ac{pca}, taking into account the bias of neural layers, and provide a deep analysis of the behavior of \ac{pca}-initialized networks -- both for \ac{cae} and \ac{cnn} -- with a focus on historical document images.
Kr{\"a}henb{\"u}hl et al.~\cite{krahenbuhl2015data} conducted a similar, but independent, research in which, while investigating data-dependent initialization, used \ac{pca} matrices as neural layer initial weights.
They however mainly focus on K-means initialization and do not investigate deeply \ac{pca} initialization.

\subsubsection*{Linear Discriminant Analysis  in Neural Networks:}

The link between \acp{nn} and \ac{lda} has been investigated by many authors. 
Among them, there are Webb and Lowe \cite{webb1990} who have shown that the output of hidden layers of multi-layer perceptrons are maximizing the network discriminant function, explicitly performing a non-linear transformation of the data into a space in which the classes may be more easily separated.
Demir and Ozmehmet \cite{demir2003} presented an online local learning algorithms for updating \ac{lda} features incrementally using error-correcting and the Hebbian learning rules.
Recently, Dorfer at al.\ \cite{dorfer2015} have shown how to learn linearly separable latent representations in an end-to-end fashion on a \ac{dnn}.
To the best of our knowledge, there have been no attempts to use \ac{lda} for direct \ac{nn} initialization.

\section{Mathematical formulation}
\label{sct:mathematical_formulation}

In this section we explain the general idea\footnote{Giving an exhaustive and complete explanation of the \ac{lda} algorithm is behind the scope of this paper. We are keeping the notation and the mathematical background as simple as possible by limiting ourselves to the essential information for understanding this paper. Unless stated otherwise we use the following notation: $v_i$ is the i-th element of $\bm{v}$.} of the Linear Discriminant Analysis and then give the mathematical formulation for using it both as features extractor and classifier.

\subsection{LDA in a Nutshell}

\ac{lda} seeks to reduce dimensionality while preserving as much of the class discriminatory information as possible \cite{ricardogutierrezosuna}. %
Assume we have a set of observations $X$ belonging to $C$ different classes. %
The goal of \ac{lda} is to find a linear transformation (projection) matrix $L$ that converts the set of labelled observations $X$ into another coordinate system $X' = X\cdot L$ such that the linear class separability is maximized and the variance of each class is minimized. %

\subsection{LDA vs PCA}

\begin{figure}[!t]
  \begin{center}
    \includegraphics[width=\columnwidth]{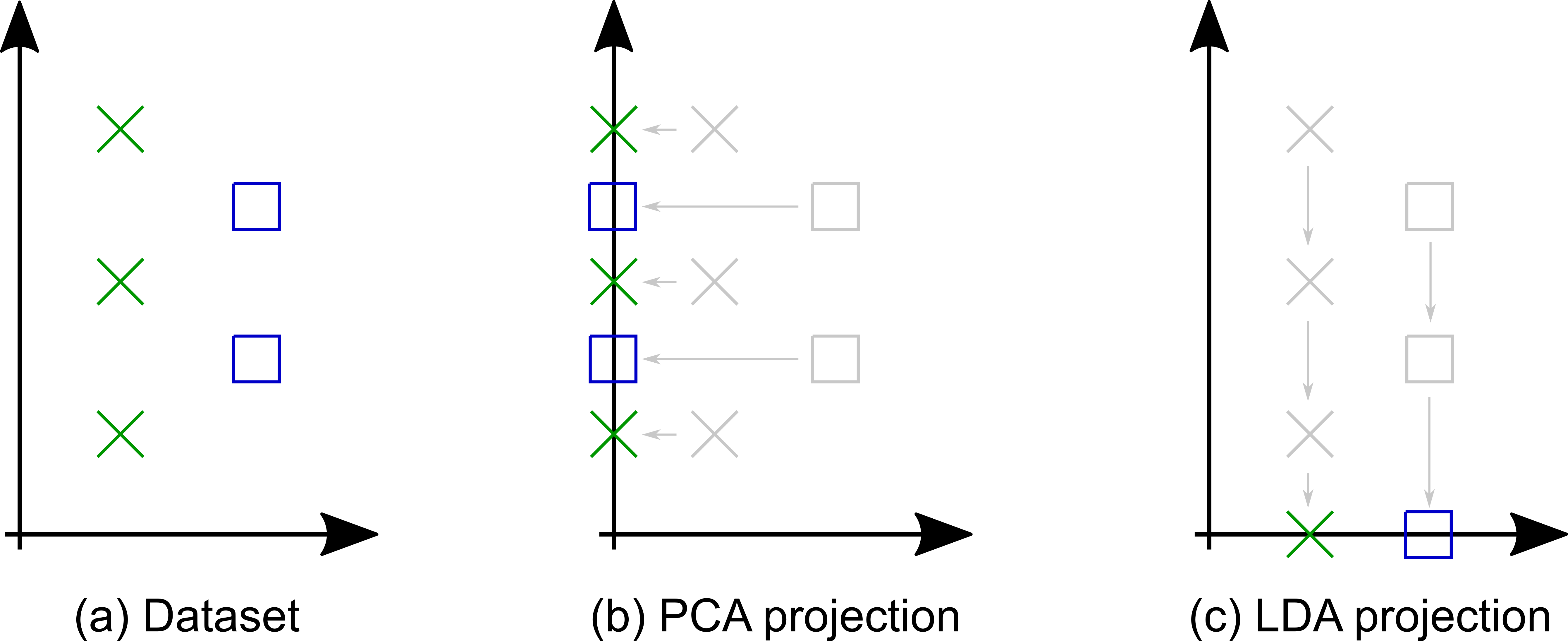}
    \caption{Example of different behaviour of LDA and PCA approaches on the same dataset. The dataset is presented in (a). In (b) the dataset projected on the basis that a PCA would chose is shown, as this maximizes the variance, regardless of the class labels (blue and green colors in this case). In (c) the dataset projected on the basis that a LDA would chose is shown, as this maximizes the class separation, in this case green from blue entries. }
    \label{fig:pca_vs_lda}
  \end{center}
\end{figure}

Both \ac{lda} and \ac{pca} are linear transformation methods and are closely related to each other \cite{martineza2001}. However, they pursue two completely different goals (see Figure~\ref{fig:pca_vs_lda}):

\begin{itemize}
\item[PCA] Looks for the directions (components) that maximize the variance in the dataset. It therefore does not need to consider class labels.
\item[LDA] Looks for the directions (components) that maximize the class separation, and for this it needs class labels.
\end{itemize}

\subsection{LDA as Feature Extractor}

In our previous work \cite{seuretalberti2017pca} we successfully initialized a \ac{nn} layer-wise with \ac{pca}. %
Here, we exploited the similarities behind the mathematical formulation of \ac{pca} and \ac{lda} to initialize a \ac{nn} layer to perform \ac{lda} space transformation. %
Recall that a standard artificial neural layer takes as input a vector $\bm{x}$, multiplies it by a weight matrix $W_{nn}$, adds a bias vector $\bm{b}$, and applies a non-linear activation function $f$ to the result to obtain the output $\bm{y}$:

\begin{equation}
\bm{y}_{nn} = f \left( W_{nn} \cdot \bm{x} + \bm{b} \right)
\label{eq:nn_layer}
\end{equation}

The \ac{lda} space transformation operation can be written in the same fashion:

\begin{equation}
\bm{y}_{lda} = W_{lda} \cdot \bm{x}
\end{equation}

Ideally, we would like to have $\bm{y}_{nn} = \bm{y}_{lda}$. %
This is not possible because of the non-linearity introduced by the function $f$. %
However, since $f$ does not change the sign of the output, we can safely apply it to the \ac{lda} as well, obtaining what we call an \textit{activated \ac{lda}}, which behaves like a neural layer:

\begin{equation}
\bm{y}_{alda} =  f \left(  W_{lda} \cdot \bm{x} \right)
\label{eq:alda_layer}
\end{equation}

At this point we can equal Equation~\ref{eq:nn_layer} to~\ref{eq:alda_layer}: 

\begin{equation*}
f \left( W_{nn} \cdot \bm{x} + \bm{b} \right) =  f \left(  W_{lda} \cdot \bm{x} \right) \\
\end{equation*}

Let $\bm{b} = \bm{0}$, then we have:

\begin{align*}
f \left( W_{nn} \cdot \bm{x} \right) &=  f \left(  W_{lda} \cdot \bm{x} \right) \\
W_{nn} \cdot \bm{x} &=  W_{lda} \cdot \bm{x} \\
W_{nn}  &=  W_{lda} \\
\end{align*}

This shows that the transformation matrix $W_{lda}$ can be used to quickly initialize the weight of a neural layer which will then perform the best possible class separation obtainable within a single layer, with regard to the layer training input data.
Note that inputs coming from previous layers might be not optimal for the task, thus fine-tuning \ac{lda}-initialized networks will improve classification accuracy of the top layer.
\par %
The rows of the matrix $W_{lda}$ are the sorted\footnote{The eigenvectors are sorted according to the corresponding eigenvalue in descending order.} eigenvectors of the squared matrix $J$ (see Equation~\ref{eq:J}). %
Typically with \ac{lda} one might take only the subset of the $|C|-1$ largest (non-orthogonal) eigenvectors (where $|C|$ denotes the number of classes), however, in this case, as the size of $W_{lda}$ has to match the one of $W_{nn}$, the number of eigenvectors taken is decided by the network architecture. %
This also implies that with a standard\footnote{There are variants of \ac{lda} which allows for extracting an arbitrary number of features \cite{wang2010} \cite{diaf2013}.} implementation of \ac{lda} we cannot have more neurons in the layer than input dimensions. %
\par %
The matrix $J$ is obtained as:

\begin{equation}
J = S_{W}^{-1} S_B
\label{eq:J}
\end{equation}

where $S_W$ and $S_B$ are the scatter matrices within-class and respectively between-classes \cite{raschka2014lda}. %
Let $\mu_c$ denote the within-class mean of class $c$, and $\mu$ denote the overall mean of all classes. The scatter matrices are then computed as follow:

\begin{equation}
S_W = \overline{N} \sum_{c \in C} \frac{1}{N_c} \sum_{\bm{x} \in c} (\bm{x} - \bm{\mu}_c)(\bm{x} - \bm{\mu}_c)^T
\label{eq:S_W}
\end{equation}

\begin{equation}
S_B = \overline{N} \sum_{c \in C}  \frac{1}{N_c}  (\bm{\mu}_c - \bm{\mu})(\bm{\mu}_c - \bm{\mu})^T
\label{eq:S_B}
\end{equation}

where $\overline{N}$ is the mean number of points per class and $N_c$ is the number of points belonging to class $c$. 

\subsection{LDA as Classifier}

Even though \ac{lda} is most used for dimensionality reduction, it can be used to directly perform data classification. %
To do so, one must compute the discriminant functions $\delta_c$ for each class $c$:

\begin{equation}
\delta_c(\bm{x}) = \bm{x}^T\Sigma_c^{-1}\bm{\mu}_c - \frac{1}{2}\bm{\mu}_c^T\Sigma_c^{-1}\bm{\mu}_c + log(\pi_c)
\label{eq:delta_c}
\end{equation}

where $\pi_c$ and $\Sigma_c$ are the prior probability \cite{friedman2001elements} and the pooled covariance matrix, for the class $c$. %
Let $n$ be the total number of observations in $X$, then the priors can be estimated as $\pi_c = N_c / n$, and $\Sigma_c$ computed as:

\begin{equation}
\Sigma_c = \frac{N_c -1}{n -|C|}  \sum_{x \in c} (\bm{x} - \bm{\mu}_c)(\bm{x} - \bm{\mu}_c)^T
\end{equation}

An observation $x$ will then be classified into class $c$ as:

\begin{equation}
c = \underset{c}{\arg\max} \enskip \delta_c(\bm{x})
\label{eq:argmax_c}
\end{equation}

The entire vector $\bm{\delta}$ can be computed in a matrix form (for all classes) given an input vector $\bm{x}$: %

\begin{equation}
\bm{\delta} = W \cdot \bm{x} +  \bm{b} 
\end{equation}

Notice the similarity to Equation~\ref{eq:nn_layer}. %
To initialize a neural layer to compute it we set the initial values of the bias $\bm{b}_c$ to the constant part of Equation~\ref{eq:delta_c}:

\begin{equation}
b_c = - \frac{1}{2}\bm{\mu}_c^T\Sigma_c^{-1}\bm{\mu}_c + log(\pi_c)
\end{equation}

and the rows of the weight matrix $W$ to be the linear part of Equation~\ref{eq:delta_c}, such that at the row $c$ we have  $\Sigma_c^{-1}\bm{\mu}_c$.

\section{Experiments Methodology}
\label{sct:experiments_methodology}

In this section we introduce the dataset,the architecture and the experimental setting used in this work, such that the results we obtained are reproducible by anyone else. %

\subsection{Dataset}
\label{ssct:dataset}

\begin{figure}[!t]
  \centering

  \subfloat[CB55, p.67v]  {\includegraphics[width=\columnwidth]{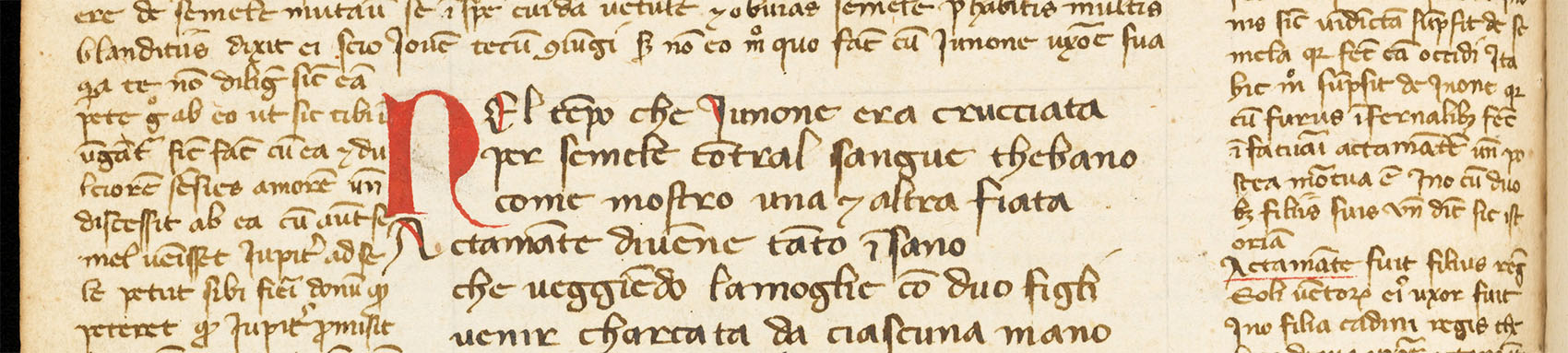}\label{fig:CB55}}
  
  \subfloat[CSG18, p.116]  {\includegraphics[width=\columnwidth]{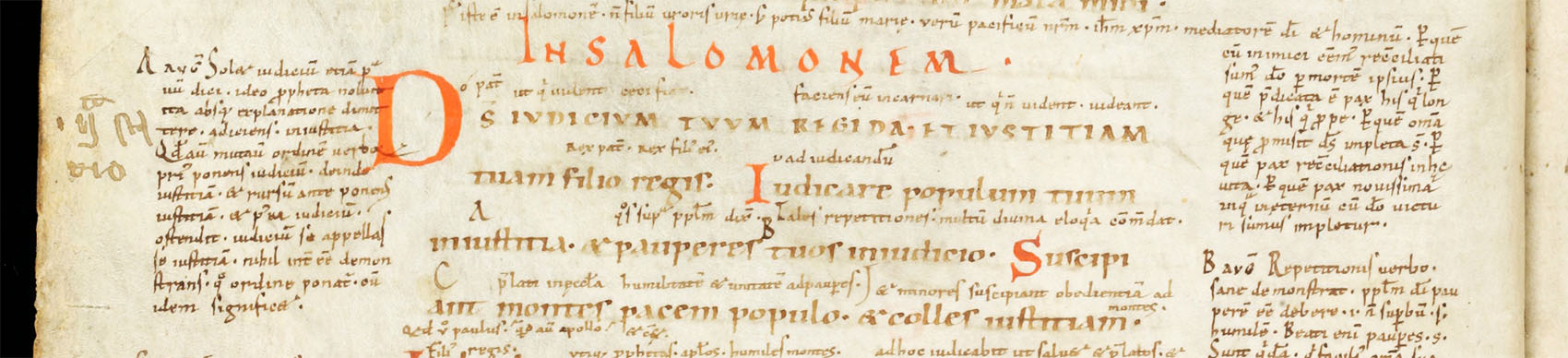}\label{fig:CSG18}}
  
  \subfloat[CSG863, p.17]  {\includegraphics[width=\columnwidth]{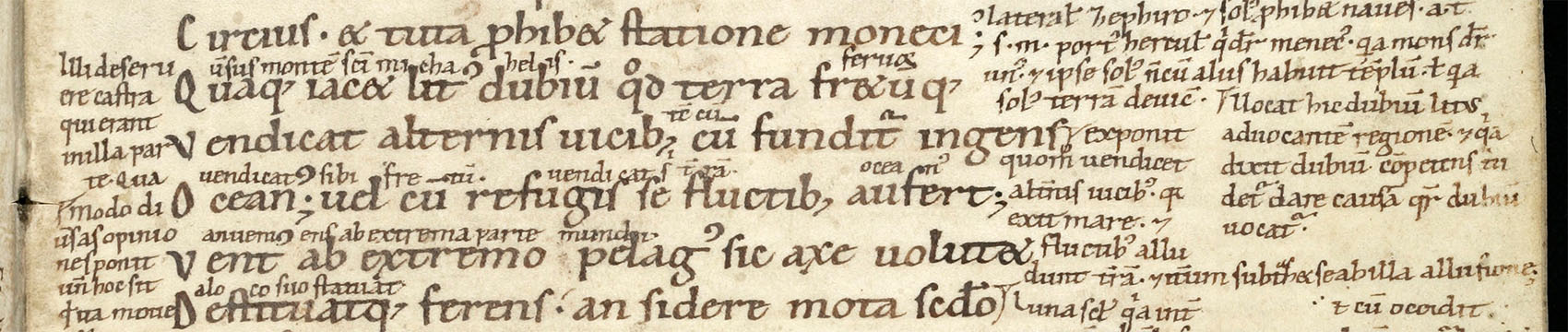}\label{fig:CSG863}}
  
  \caption{Samples of pages of the three medieval manuscripts in DIVA-HisDB.}
  \label{fig:dataset}
\end{figure}

To conduct our experiments we used the DIVA-HisDB dataset\cite{simistira_2016_diva}, which is a collection of three medieval manuscripts (CB55\footnote{Cologny-Geneve, Fondation Martin Bodmer, Cod. Bodmer 55.}, CSG18\footnote{St. Gallen, Stiftsbibliothek, Cod. Sang. 18, codicological unit 4.} and CSG863\footnote{St. Gallen, Stiftsbibliothek, Cod. Sang. 863.}) with a particularly complex layout (see Figure~\ref{fig:dataset}). %
\par %
The dataset consists of 150 pages in total and it is publicly available\footnote{\url{http://diuf.unifr.ch/hisdoc/diva-hisdb}}. %
In particular, there are 20 pages/manuscript for training, 10 pages/manuscript for validation and  10 test pages. %
There are four classes in total: background, comment, decoration and text. %
\par %
The images are in JPG format, scanned at 600 dpi, RGB color. %
The ground truth of the database is available both at pixel level and in the PAGE XML~\cite{pletschacher2010page} format. %
We chose this dataset as it as been recently used for an ICDAR competition on layout analysis \cite{simistira_2017_competition}. %
To perform our experiments we used a scaled version of factor $1/10$ in order to reduce significantly the computation time. %

\subsection{Network Architecture}
\label{ssct:network_architecture}

When designing a \ac{nn}, there is no trivial way to determine the optimal architecture hyper-parameters \cite{wang1994optimal}\cite{kavzoglu1999determining} and often the approach is finding them by trial and error (validation). %
In this work we are not interested into finding the best performing network topology as we are comparing the results of different initialization techniques on the same network. %
Therefore we used similar parameters to our previous work on this dataset \cite{simistira_2016_diva}. %
The parameters presented in the following table define the \ac{cnn} architecture for what concerns the number of layers, size of the input patches with their respective offsets\footnote{Some literature refer to offset as \textit{stride}.} and number of hidden layers. 
Each layer has a Soft-Sign activation function and the total input patch covered by the \ac{cnn} is $23 \times 23$ pixels. %
On top of these feature extraction layers we put a single classification layer with 4 neurons: one for each class in the dataset. %

\begin{table}[H]
\centering
\begin{tabular}{r|ccc}
layer           & 1         & 2          & 3         \\ \hline
patch size     & $5\times5$ & $3\times3$ & $3\times3$ \\
offset         & $3\times3$ & $2\times2$ &            \\  \hline
hidden neurons & 24         & 48         & 72        
\end{tabular}
\end{table}

\subsection{Experimental Setup}
\label{ssct:experimental_setup}

In order to investigate the effectiveness of our novel initialization method, we measure the performances of the same network (see Section~\ref{ssct:network_architecture}) once initialized with \ac{lda} and once with random weights. %
We evaluate the network for the task of layout analysis at pixel level in a multi-class, single-label setting\footnote{This means that a pixel belongs to one class only, but it could be one of many different classes.}. %
\subsubsection*{Initializing with \ac{lda}}
\begin{figure}[!t]
  \centering
  \includegraphics[width=\columnwidth]{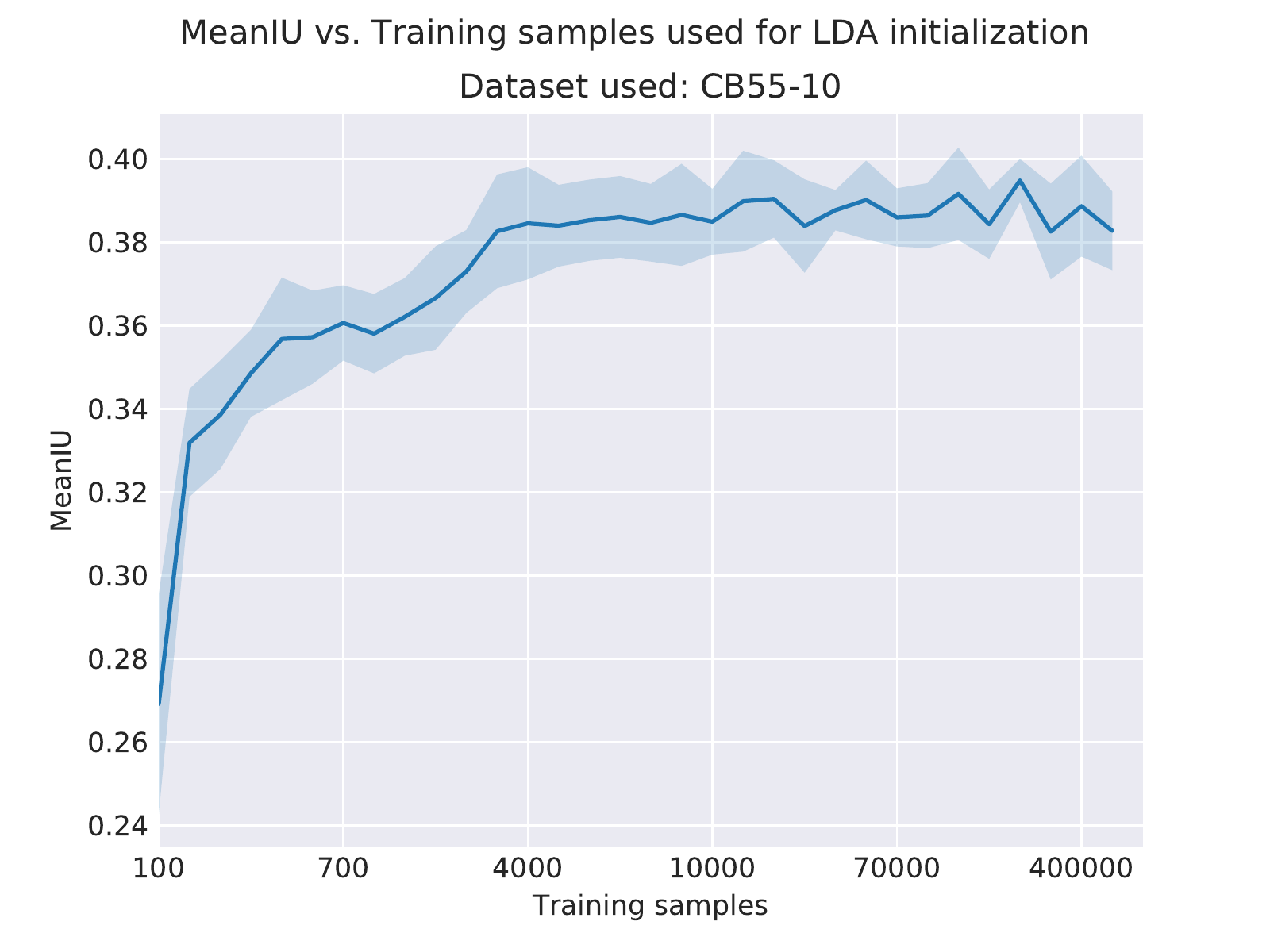}
  \caption{Stability of the \ac{lda} initialization in respect to the number of training samples used to compute it on CB55. The shaded area represents the variance whereas the tick line is the mean value, computed over 10 runs. Mind that the x-axis is not linear. Observe how as of 40k samples it seems to have reached the maximum performance.}
  \label{fig:LDAvsSamples}
\end{figure}

First we evaluate the stability of the \ac{lda} initialization in respect to the number of training samples $k$ used to compute it (see Figure~\ref{fig:LDAvsSamples}) . %
After validating this hyper-parameter, we will use it for all other experiments. %
\par %
When initializing a multi layer network with \ac{lda} we start by computing \ac{lda} on $k$ raw input patches and use the transformation matrix to initialize the first layer. %
We then proceed to apply a forward pass with the first layer to all $k$ raw input patches and we use the output to compute again \ac{lda} such that we can use the new transformation matrix to initialize the second layer. %
This procedure is then repeated until the last layer is initialized. %
At this point, we add a classification layer that we will initialize in the same fashion as the others, but with the linear discriminant matrix (see Section~\ref{sct:mathematical_formulation}) rather than with the transformation matrix. %
The whole procedure takes less than two minutes with our experimental setting.
\subsubsection*{Initializing with random weights}
For the random initialization we trivially set the weights matrices to be randomly distributed in the range $\left[\frac{1}{-\sqrt{n}}, \frac{1}{\sqrt{n}}\right]$, where $n$ is the number of inputs of the neuron of the layer being initialized. %
\subsubsection*{Training phase}
Once the networks are initialized (both \ac{lda} and random) we test their performance on the test set already and again after each training epoch. %
We then train them for $100$ epochs (where one epoch corresponds to $100K$ training samples) with mini-batches of size $4096$. %
We optimize using standard SGD (Stochastic Gradient Descent) with learning rate $0.01$. %
In order to reduce the role randomness play in the experiments, in a single run we show the same input patches to both an \ac{lda} and a random network --- so that pair-wise they see the same input --- and the final results are computed by averaging 10 runs. 
\subsubsection*{Evaluation metric}
The evaluating metric chosen is the mean Intersection over Union because is much stricter than accuracy and especially is not class-size biased \cite{alberti2017evaluation}. %
We measure it with an open-source\footnote{Available at \url{https://github.com/DIVA-DIA/LayoutAnalysisEvaluator}.} tool. %

\section{Features Visualization}
\label{sct:features}

\begin{figure}[!t]
  \centering
  
  $\vcenter{\hbox{\subfloat[First layer features (LDA)]%
  {\includegraphics[width=.45\columnwidth]{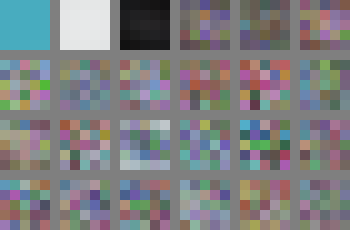}%
  \label{fig:features_LDA_1}}}}$
  \hfil
  $\vcenter{\hbox{\subfloat[First layer features (RND)]%
  {\includegraphics[width=.45\columnwidth]{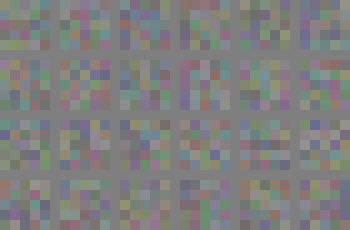}%
  \label{fig:features_RND_1}}}}$
  
  $\vcenter{\hbox{\subfloat[Second layer features (LDA)]%
  {\includegraphics[width=.45\columnwidth]{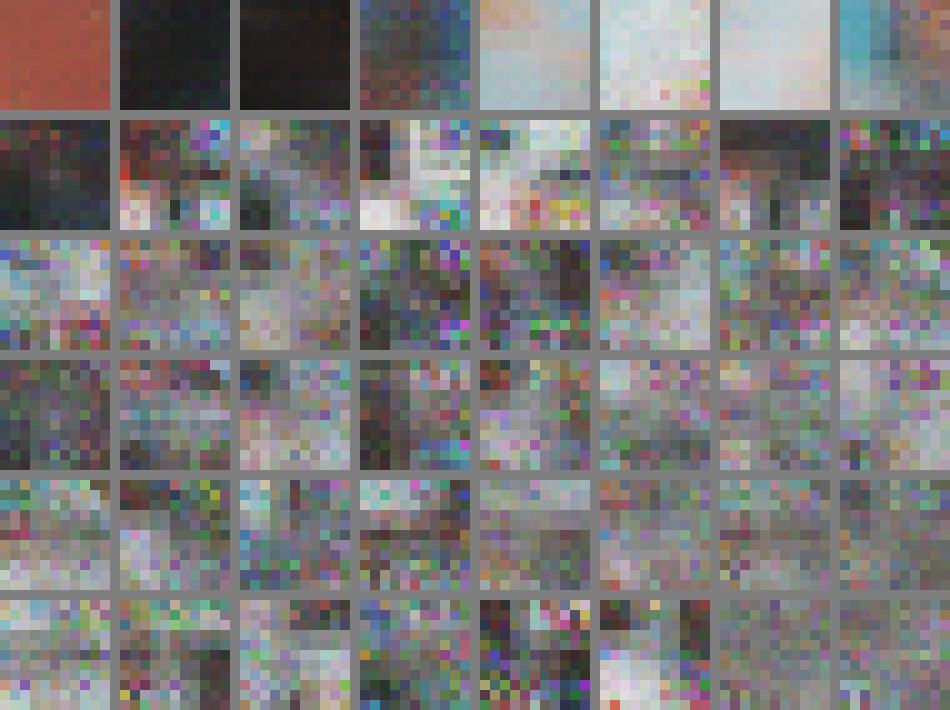}%
  \label{fig:features_LDA_2}}}}$
  \hfil
  $\vcenter{\hbox{\subfloat[Third layer features (LDA)]%
  {\includegraphics[width=.45\columnwidth]{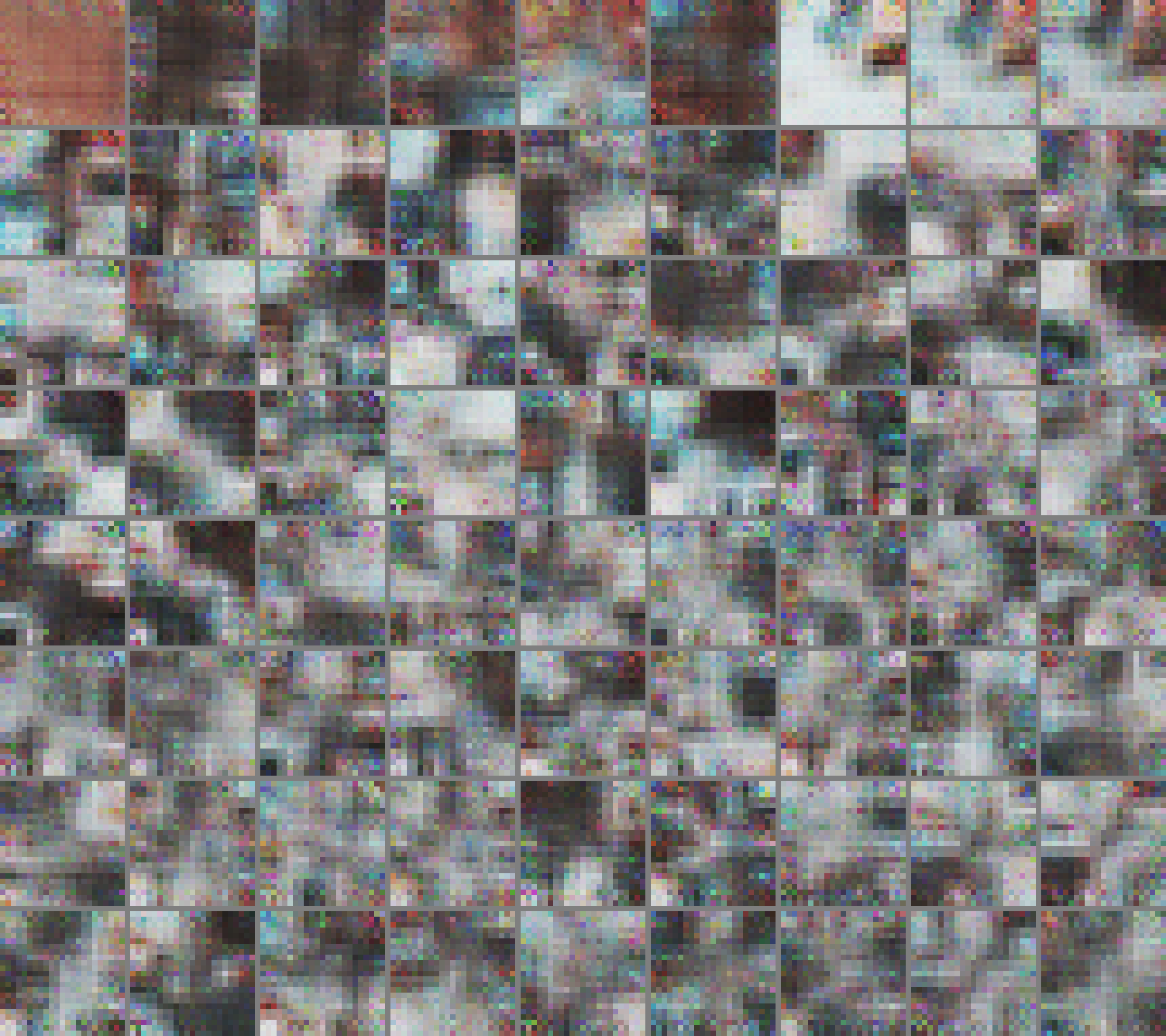}%
  \label{fig:features_LDA_3}}}}$
  
  \caption{Features of the first three layers of the network initialized with \ac{lda} and of the first layer of a network randomly initialized, for the CSG863 manuscript. Notice how the \ac{lda} ones are much more colorful than the random ones. Moreover, in (a) one can see how,  as expected, there are 3 meaningful features, which are exactly as many as the number of classes minus one (see details in Sections~\ref{sct:mathematical_formulation} and~\ref{sct:features}).}
  \label{fig:features}
\end{figure}

In this section we show and briefly discuss the features visualization of the \ac{cnn} initialized with \ac{lda}. %
In Figure~\ref{fig:features} are shown the features of the first three layers of the network initialized with \ac{lda} and of the first layer of a network randomly initialized, for the CSG863 manuscript. %
\par
\begin{figure}[!t]
  \centering
  \subfloat[\ac{lda} init network before training]%
  {\includegraphics[width=.47\columnwidth]{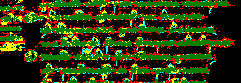}\label{subfig:lda_visualization_before}}
  \hfil
  \subfloat[\ac{lda} init network after training]%
  {\includegraphics[width=.47\columnwidth]{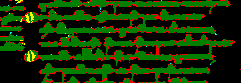}\label{subfig:lda_visualization_after}}

  \subfloat[RND init network before training]%
  {\includegraphics[width=.47\columnwidth]{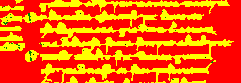}\label{subfig:rnd_visualization_before}}
  \hfil
  \subfloat[RND init network after training]%
  {\includegraphics[width=.47\columnwidth]{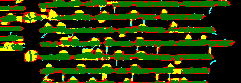}\label{subfig:rnd_visualization_after}} 

  \caption{Visualization of the classification performances of both \ac{lda} and random (RND) initialized network, before and after training for a small patch of CB55. Green and black denotes correctly classified foreground and background pixels, blue and red are false negative respectively false positive foreground pixels and yellow is used for wrong foreground class e.g text instead of comment.}
  \label{fig:visualization}
\end{figure}

 %
Without surprise, the features produced by the random initialization are not visually appealing as they are very close to being just gray with noise. %
In fact, they are not representing something meaningful at all\footnote{For this reason, we displayed only the first layer in Figure~\ref{fig:visualization}: images of layer two and three were not conveying additional information.}. %
\par %
On the other hand, those produced by the \ac{lda} initialization are a completely different story. %
Notice how on the first layer (Figure~\ref{fig:features_LDA_1}), there are 3 meaningful features, which are exactly as many as the number of classes minus one (see details in Section~\ref{sct:mathematical_formulation}). %
We expected the first three features to be ''mono-color`` and much different than the others, as we know standard \ac{lda} typically projects the points in a sub-dimensional space of size $|C|-1$, where $|C|$ is the number of classes (see details in Section~\ref{sct:mathematical_formulation}). %
Moreover, the other 21 features are yes, looking like the random ones, but are much more colorful. %
This means that their values are further away from zero. %
\par %
Regarding the second and third layer (Figures~\ref{fig:features_LDA_2} and \ref{fig:features_LDA_3}), as convolution is involved is difficult to interpret their visualization in an intuitive way. %
We can, however, observe how also in the second layer the first three features are significantly different than the other ones and how this is not entirely true anymore in the third layer. %

\section{Results Analysis}
\label{sct:results}

%
%
%





%
%

\begin{figure}[!tb]
  \centering
  \subfloat[Mean IU during training for CB55]%
  {\includegraphics[width=.97\columnwidth]{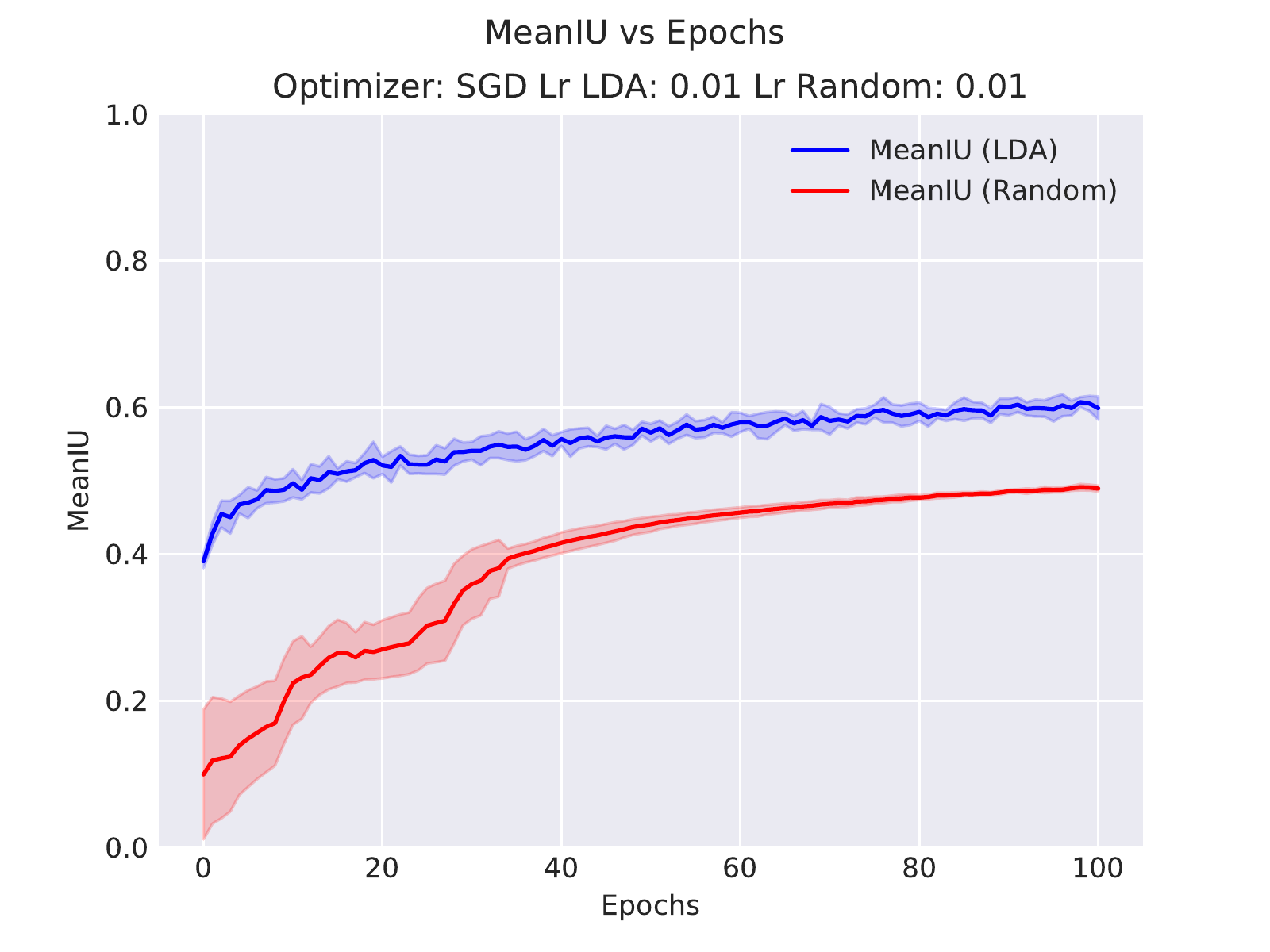}\label{subfig:r1}}
  
  \subfloat[Mean IU during training for CS18]%
  {\includegraphics[width=.47\columnwidth]{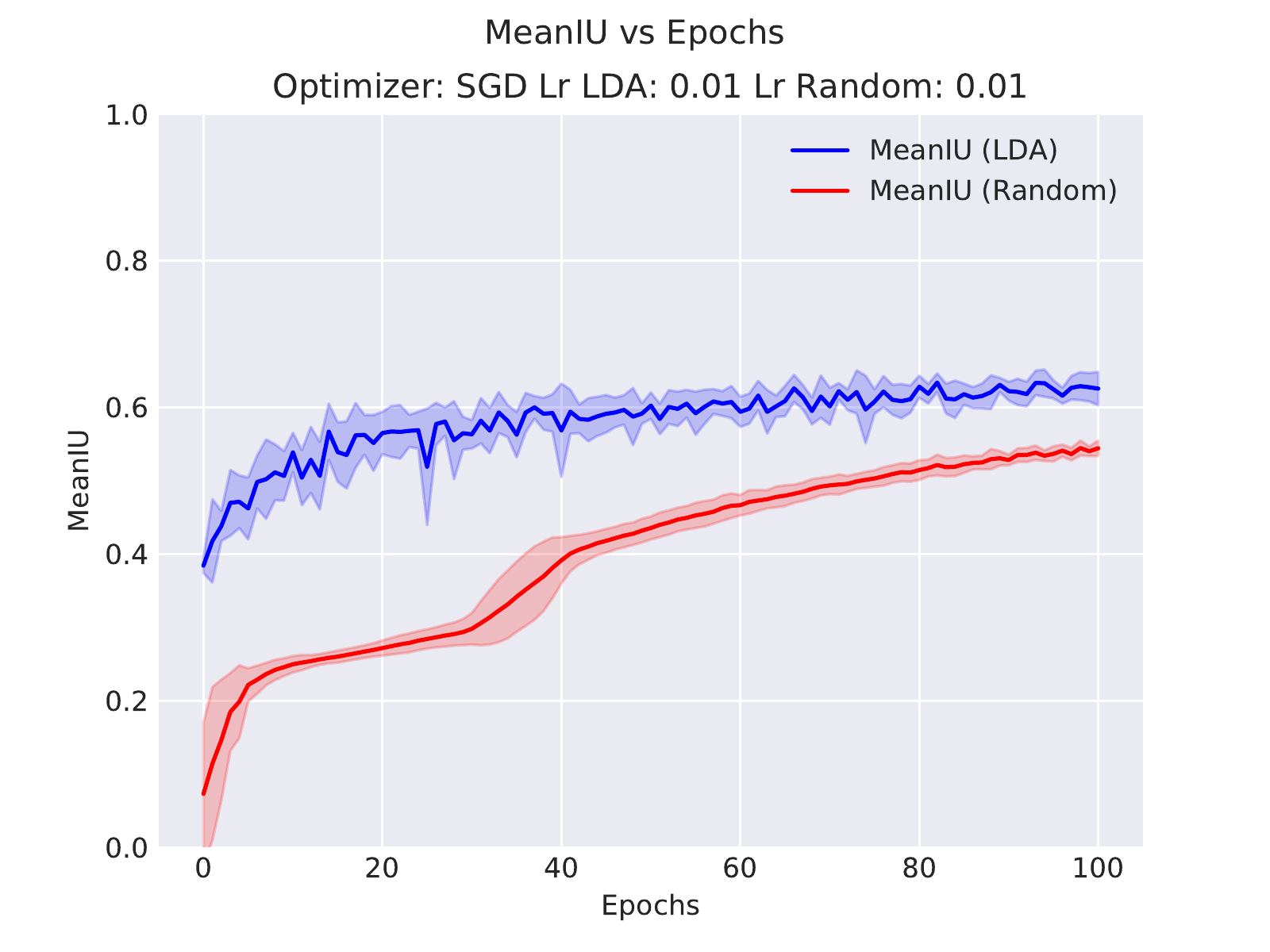}\label{subfig:r2}}
  \hfill
  \subfloat[Mean IU during training for CS863]%
  {\includegraphics[width=.47\columnwidth]{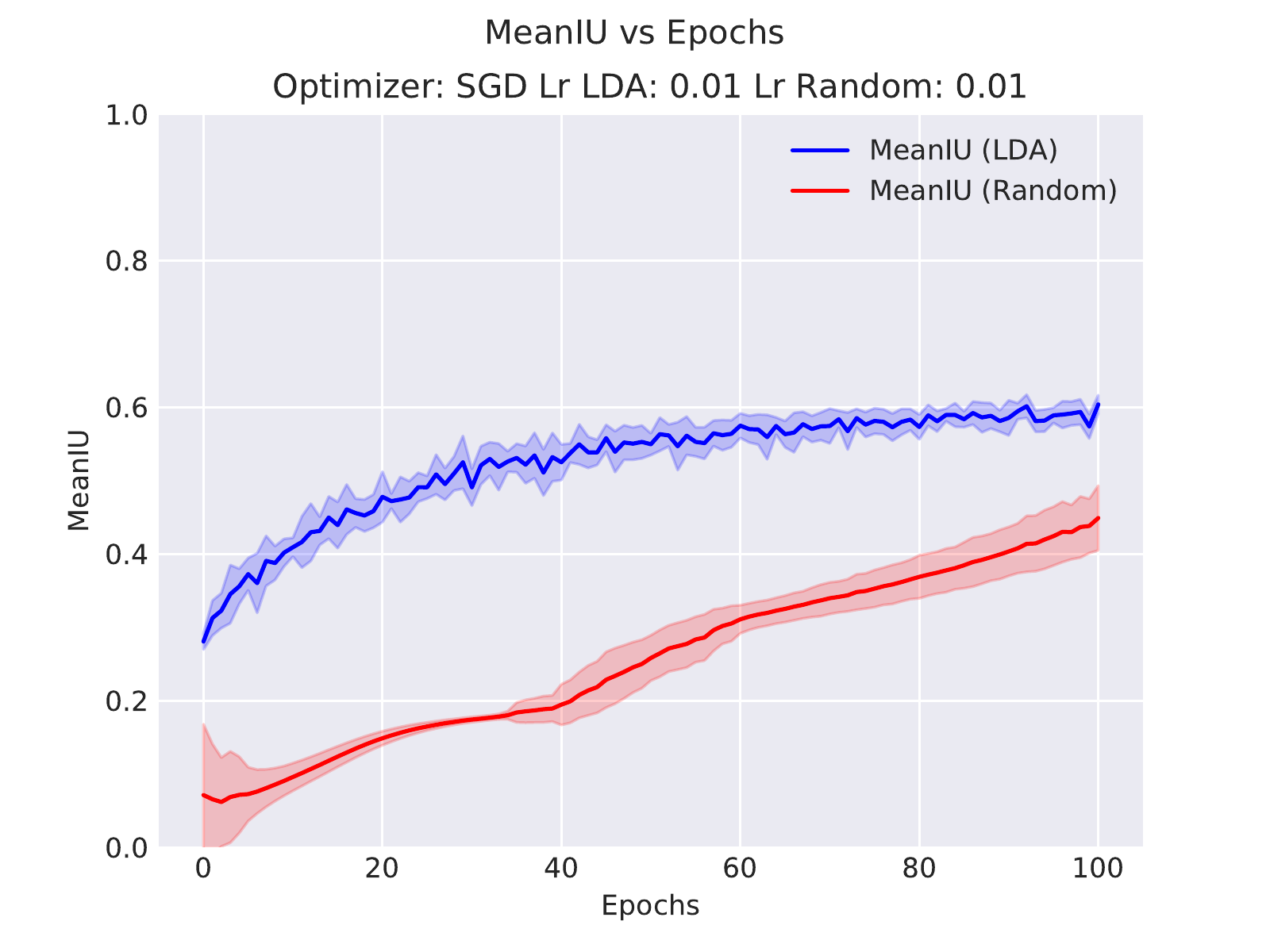}\label{subfig:r3}} 

  \caption{
    Mean performances of the networks initialized with \ac{lda} (blue) and random weights (red) evaluated during the training.
    \ac{lda} networks are significantly better than the random counterpart: they start off at a very high IU and converge to a better local minima.
  }
  \label{fig:3results}
\end{figure}

We measured the mean IU of networks during their training phase, evaluating them after each epoch~--~Figure~\ref{fig:3results} shows their performances.
The \ac{lda} initialization is very stable as all networks started with almost the same mean IU.
The random initialization however leads to a very high variance of the classification quality at the beginning of the training.

We can also note that the curves of the \ac{lda}-initialized networks have very similar shapes for all three manuscripts, thus their behavior can be considered as rather predictable.
Contrariwise, the random initialization leads to three different curve shapes, one per manuscript, so we cannot predict how randomly-initialized networks would behave on other manuscripts.

The \ac{lda} initialization has two capital advantages over the random one.
First, initial mean IU clearly outperforms randomly-initialized networks, as shown in Table~\ref{tab:LDAvsRND}.
The table also includes the percentage of correctly classified pixels, a measurement less punitive than the mean IU but which sheds light from another angle on the advantages of \ac{lda} initialization.
Second, the \ac{lda} initialization leads quickly to much better local minima.
In the case of CS863, none of the 10 random networks has finished converging after 100 epochs while \ac{lda}-initialized networks have almost finished converging after 60 epochs.

These advantages can be explained by looking at the features obtained by \ac{lda}-initialization shown in Figure~\ref{fig:features}. There are useful structures for some of the filters in all three layers of the network before starting the training, thus less weight adaptations are needed.

In the case of CB55 and CS18, randomly-initialized networks seem to all find similar solutions, and end with very low mean IU variance. Observing only these results, one could think this is the best that can be obtained with the network topology we used, yet the \ac{lda} initialization proves this assertion wrong.

\begin{table}[!t]
\centering
\begin{tabular}{l|ccc|c|c}
                &   \multicolumn{4}{c|}{Mean IU}            & Accuracy  \\
model           &   CB55    &   CSG18   &   CSG863  &   AvG &   AvG     \\  \hline
LDA(init)       &   0.39       &   0.38       &   0.28       &   0.35   &   0.75       \\
RND(init)       &   0.09       &    0.07      &   0.07       &   0.07   &   0.20      \\  \hline
LDA(trained)    &   0.59       &   0.62       &   0.60       &   0.60   &   0.93       \\
RND(trained)    &   0.48       &   0.54       &   0.44       &   0.48   &   0.88       \\
\end{tabular}
\caption{Results obtained with both \ac{lda} and random initialized networks on all manuscripts. The most-right column is the average accuracy over the whole dataset. We added it to emphasize the strictness of the Mean IU metric, which is always much lower than the accuracy, as it is not class-size biased.}
\label{tab:LDAvsRND}
\end{table}


\section{Conclusion and Outlook}

In this paper, we have investigated a new approach for initializing \ac{dnn} using \ac{lda}.
We show that such initialization is more stable, converge faster and to better performances than the random weights initialization.
This leads to significantly shorter training time for layout analysis tasks at the cost of an initialization time that can be considered as negligible.


\par %
This study has been conducted only on relatively small \ac{cnn}, so the generality of the aforementioned findings should be investigated for deeper networks. %
Also, as the focus was not achieving high level of accuracy, the design of our test is kept small and simple. %
Consequently, the results obtained should not be compared to state of art ones. %

\par %
As future work, we intend to study the joint use of multiple statistical methods (such as \ac{pca} and \ac{lda}) to initialize a much deeper \ac{dnn} and to extend the performances test to other classification tasks (e.g image recognition, digit recognition).





Finally, we believe that a good network initialization might be a solution to reduce the training time of \ac{dnn} significantly. %

\section*{Acknowledgment}
The work presented in this paper has been partially supported by the HisDoc III project funded by the Swiss National Science Foundation with the grant number $205120$\textunderscore$169618$.

\bibliographystyle{ACM-Reference-Format}
\bibliography{biblio} 

\end{document}